\documentclass[a4paper,twoside]{article}

\usepackage{epsfig}
\usepackage{subcaption}
\usepackage{calc}
\usepackage{amssymb}
\usepackage{amstext}
\usepackage{amsmath}
\usepackage{amsthm}
\usepackage{multicol}
\usepackage{pslatex}
\usepackage{apalike}
\usepackage[bottom]{footmisc}

\usepackage{multirow}
\usepackage{adjustbox}
\usepackage{xcolor}
\usepackage{pgfplots}
\pgfplotsset{compat=1.17}
\usepackage{enumitem}

\usepackage{SCITEPRESS}     

\begin{document}

\title{Linking data separation, visual separation, and classifier performance using pseudo-labeling by contrastive learning}

\author{\authorname{Bárbara Caroline Benato\sup{1}\orcidAuthor{0000-0003-0806-3607}, Alexandre Xavier Falcão\sup{1}\orcidAuthor{0000-0002-2914-5380}, and Alexandru-Cristian Telea\sup{2}\orcidAuthor{0000-0003-0750-0502}}
\affiliation{\sup{1}Laboratory of Image Data Science, Institute of Computing,\\ University of Campinas, Campinas, Brazil}
\affiliation{\sup{2}Department of Information and Computing Sciences, Faculty of Science, \\Utrecht University, Utrecht, The Netherlands}
\email{\{barbara.benato, afalcao\}@ic.unicamp.br, a.c.telea@uu.nl}
}



\abstract{Lacking supervised data is an issue while training deep neural networks (DNNs), mainly when considering medical and biological data where supervision is expensive. Recently, \emph{Embedded Pseudo-Labeling} (EPL) addressed this problem by using a non-linear projection (t-SNE) from a feature space of the DNN to a 2D space, followed by semi-supervised label propagation using a connectivity-based method (OPFSemi). We argue that the performance of the final classifier depends on the data separation present in the latent space and visual separation present in the projection. We address this by first proposing to use contrastive learning to produce the latent space for EPL by two methods (SimCLR and SupCon) and by their combination, and secondly by showing, via an extensive set of experiments, the aforementioned correlations between data separation, visual separation, and classifier performance. We demonstrate our results by the classification of five real-world challenging image datasets of human intestinal parasites with only $1\%$ supervised samples. 
}

\onecolumn \maketitle \normalsize \setcounter{footnote}{0} \vfill

\section{\uppercase{Introduction}}
\label{sec:introduction}
While supervised learning has achieved great success, using datasets with either (i) few data points or (ii) few supervised, \emph{i.e.} labeled, points, is fundamentally hard, and especially critical in \emph{e.g.} medical contexts where obtaining (labeled) points is expensive. For (i), methods such as few-shot learning\,\cite{Sung:2018,Sun:2017}, transfer-learning\,\cite{imagenet:2015}, and data augmentation have been used to increase the sample count. For (ii), solutions include semi-supervised learning\,\cite{Iscen:2019,Wu:2018}, pseudo-labeling\,\cite{Lee:2013,Jing:2020}, and meta-learning~\cite{Pham:2021:CVPR}. 

Pseudo-labeling, also called self-training, takes a training set with few supervised and many unsupervised samples and assigns pseudo-labels to the latter samples -- a process known as data annotation -- and re-trains the model with all (pseudo)labeled samples. Yet, as the name suggests, pseudo-labels are not perfect, as they are \emph{extrapolated} from actual labels, which can affect training performance\,\cite{BenatoSibgrapi:2018,Arazo:2020}. Also, pseudo-labeling methods still require training and validation sets with thousands of supervised samples per class to yield reasonable results\,\cite{Miyato:2018,Jing:2020,Pham:2021:CVPR}. 

Both pseudo-labeling, and broader, the success of training a classifier, depend on a key aspect -- how easy is the data \emph{separable} into different groups of similar points. Projections, or dimensionality reduction methods, are well known techniques that aim to achieve precisely this\,\cite{nonato18,espadoto19}. Two key observations were made in this respect (discussed in detail in Sec.~\ref{sec:relatedworks}):

\begin{enumerate}
    \item[O1] Visual separability (VS) in a projection mimics the data separability (DS) in the high dimensional space;
    \item[O2] Data separability (DS) is key to achieving high classifier performance (CP);
\end{enumerate}

These observations have been used in several directions, \emph{e.g.}, using projections to assess DS (VS$\rightarrow$DS)\,\cite{Maaten:review}; using projections to find which samples get misclassified (VS$\rightarrow$CP)\,\cite{nonato18}; increasing DS to get easier-to-interpret projections (DS$\rightarrow$VS)\,\cite{Kim:2022}; using projections to assess classification difficulty (VS$\rightarrow$CP)\,\cite{RauberInfVis2017,Rauber:2016}; and using projections to build better classifiers (VS$\rightarrow$CP)\,\cite{BenatoSibgrapi:2018,Benato:2020:CIARP}. However, to our knowledge, no work so far has explored the relationship between DS, VS, \emph{and} CP in the context of using pseudo-labeling for machine learning (ML). 

We address the above by studying how to generate a high DS using \emph{contrastive learning} approaches which have shown state-of-the-art results\,\cite{Chen:2020,Grill:2020,he:2020:Moco,Khosla:2020} and have surpassed results of (self-and-semi-) supervised methods and even known supervised loss functions such as cross-entropy\,\cite{Chen:2020}. We compare two contrastive learning models (SimCLR\,\cite{Chen:2020} and SupCon\,\cite{Khosla:2020}) and propose a hybrid approach that combines both. We evaluate DS by measuring CP for a classifier trained with only $1\%$ supervised samples. Then, we evaluate VS fed with the encoder's output of our trained contrastive models. Lastly, we investigate CP by using our above pseudo-labeling to train a deep neural network. We perform all our experiments in the context of a challenging medical application (classifying human intestinal parasites in microscopy images).

Our main contributions are as follows:

\begin{enumerate}
    \item[C1:] We use contrastive learning to reach high DS;
    \item[C2:] We show that projections constructed from contrastive learning methods (with good DS) lead to a good VS between different classes;  
    \item[C3:] We train classifiers with pseudo-labels generated via good-VS projections to achieve a high CP.
\end{enumerate}

Jointly taken, our work brings more evidence that links the observations O1 and O2 mentioned above, \emph{i.e.}, that VS, DS, and CP are strongly correlated and that this correlation, and 2D projections of high-dimensional data, can be effectively \emph{used} to build higher-CP classifiers for the challenging case of training-sets having very few supervised (labeled) points.



\section{\uppercase{Related work}}
\label{sec:relatedworks}
%

\noindent\textbf{Self-supervised learning.} Self-supervised contrastive methods in representation learning have been the choice for learning representations without using any labels\,\cite{Chen:2020,Grill:2020,he:2020:Moco,Khosla:2020}. Such methods work by using a so-called 
\emph{contrastive loss} to pull similar pairs of samples closer while pushing apart dissimilar pairs. To select (dis)similar samples without using label information, one can generate multiple views of the data via transformations. For image data, SimCLR\,\cite{Chen:2020} used transformations such as cropping, Gaussian blur, color jittering, and grayscale bias. MoCo\,\cite{he:2020:Moco} explored a momentum contrast approach to learn a representation from a progressing encoder while increasing the number of dissimilar samples. BYOL\,\cite{Grill:2020} used only augmentations from similar examples. SimCLR has shown significant advances in (self-and-semi-) supervised alearning and achieved a new record for image classification with few labeled data. Supervised contrastive learning (SupCon)\,\cite{Khosla:2020} generalized both SimCLR and N-pair losses and was proven to be closely related to triplet loss. SupCon surpasses cross-entropy, margin classifiers, and other self-supervised contrastive learning techniques.

\smallskip
\noindent\textbf{Pseudo-labeling.} An alternative to building accurate and large training sets is to \emph{propagate} labels from a few supervised samples to a large set of unsupervised ones by creating pseudo-labels. \cite{Lee:2013} trained a neural network with $100$ to $3000$ supervised images and then assigned the class with maximum predicted probability to the remaining unsupervised ones. The network is then fine-tuned using both true and pseudo-labels to yield the final model. Yet, this method requires a validation set with over $1000$ supervised images to optimize hyperparameters. The same issue happens for other pseudo-labeling strategies that need a validation set\,\cite{Miyato:2018,Jing:2020,Pham:2021:CVPR}.

\smallskip
\noindent\textbf{Structure in (embedded) data.}
\emph{Data structure}, also called data separability (DS) is an accepted, albeit not formally, defined term in ML. Simply put, for a dataset $D =\{\mathbf{x}_i | \mathbf{x}_i \in \mathbb{R}^n\}$, DS refers to the presence of \emph{groups} of points which are similar and also separated from other point groups. DS is essential in ML, especially classification. Obviously, the stronger DS is, the easier is to build a classifier that separates points belonging to the various groups with high classifier performance (CP). CP can be measured by many metrics, \emph{e.g.}, accuracy, F1 score, or AUROC\,\cite{Hossin:2015}. Indeed, if different-class points are not separated via their features (coordinates in $\mathbb{R}^n$), then no (or poor) classification (CP) is possible. 

Projections, or Dimensionality Reduction (DR) methods, take a dataset $D$ and produce a scatterplot, or embedding of $D$, $P(D) = \{\mathbf{y}_i = P(\mathbf{x}_i) | \mathbf{y}_i \in \mathbb{R}^q\}$, where typically $q \in \{2,3\}$. The aim is that the \emph{visual structure}, also called visual separability (VS) in $P(D)$, literally seen in terms of point clusters separated by whitespace, mimics DS. Many methods have been proposed for $P$, with accompanying metrics to gauge how much VS captures DS\,\cite{nonato18,espadoto19}. 

Relations between VS, DS, and CP have been \emph{partially} explored. \cite{RauberInfVis2017} used the VS of a t-SNE\,\cite{MaatenJMLR:14} projection to gauge the difficulty of a classification task (CP). They found that VS and CS are positively correlated when VS is medium to high but could not infer actionable insights for low-VS projections. Also, they did not address the task of \emph{building} higher-CP classifiers using t-SNE.
In a related vein, \cite{rodrigues:2019} used the VS in projections to construct so-called decision boundary maps to interpret classification performance (CP) but did not actually use these to improve classifiers.
\cite{Kim:2022,sdrnnp} showed that one can improve VS by increasing DS, the latter being done by mean shift\,\cite{comaniciu}. However, their aim was to generate easier-to-interpret projections and not use these to build higher-CP classifiers. Moreover, their approach actually changed the input data in ways not easy to control, which raises question as to the interpretability of the resulting projections. Next, \cite{BenatoSibgrapi:2018,Benato:2021:PR} used the VS of t-SNE projections to create pseudo-labels and train higher-CP classifiers from them. They showed that label propagation in the 2D projection space can lead to higher-CP classifiers than when propagating labels in the data space. Yet, they did not study how correlations between DS and VS can affect CP. 

\smallskip
\noindent\textbf{Embedded Pseudo-Labeling (EPL).} The abovementioned topics of pseudo-labeling and VS-CP correlation were connected recently by \emph{Embedded Pseudo-Labeling} (EPL)~\cite{BenatoSibgrapi:2018}, a method proposed to increase the number of labeled samples from only dozens of supervised samples, without needing validation sets with more supervised samples. To do this, EPL projects to 2D the latent feature space extracted from a deep neural network (DNN) using autoencoders\,\cite{Benato:2021:PR} and pre-trained architectures\,\cite{Benato:2020:CIARP}. Pseudo-labels are next propagated in the 2D projection from supervised to unsupervised samples using the OPFSemi\,\cite{Amorim:2016} method. However, the success of EPL strongly depends on the VS in the projection space.

\section{\uppercase{Proposed pipeline}}
\label{sec_pipeline}
\begin{figure}[htb!]
    \centering
    \includegraphics[width=\linewidth]{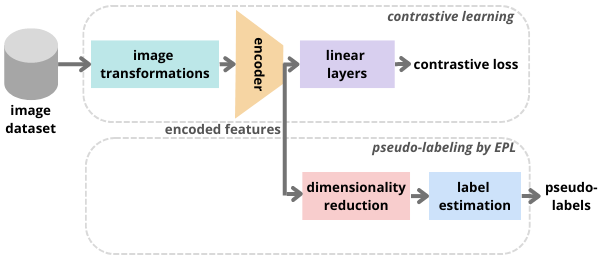}
    \caption{We train a model from image transformations of the original data with a contrastive learning loss. Next, we project the latent features from the encoder's output to 2D and pseudo-label the resulting points. Finally, we use these pseudo-labels to train a classifier.}
    \label{pipeline_fig1}
\end{figure}

Following the above, we propose to improve DS in the feature space that EPL takes as input by using two \emph{contrastive learning} models (SimCLR\,\cite{Chen:2020} and SupCon\,\cite{Khosla:2020}, used both separately and combined) and without using ground-truth labels. The feature space to input in EPL comes from the encoder's output from these contrastive models. During the process, outlined in Fig.~\ref{pipeline_fig1}, we test our three claims (Sec.~\ref{sec:introduction}), i.e., that DS has improved (C1); that this has led to an improved VS in the 2D projections used by EPL (C2); and finally that the generated pseudo-labels by EPL can be used to train a classifier with high CP (C3). Our method is detailed next. 

\subsection{Contrastive learning}
\label{sec:contr_learn}

We generate the latent space to be used by EPL (Fig.~\ref{pipeline_fig1}, top box) in three different ways: (a) from the many unsupervised samples available by using SimCLR\,\cite{Chen:2020}; (b) using our $1\%$ supervised samples with SupCon\,\cite{Khosla:2020}; and (c) by combining the SimCLR and SupCon methods. 

\subsection{Pseudo-labeling by EPL}
\label{sec:pseudolabeling}
Both SimCLR and SupCon use ResNet-18\,\cite{he:2016} as encoder. We reduce the output of ResNet-18 (hundreds of dimensions)
to 2D using t-SNE (Fig.~\ref{pipeline_fig1}, middle box). This is similar to EPL, which has shown that propagating pseudo-labels in this 2D space creates large labeled training-sets that lead to high-CP classifiers\,\cite{Benato:2020:CIARP,Benato:2021:Sibgrapi}.
We use the 2D projection to propagate the (few) true labels to all unsupervised points as in EPL. That is, we use OPFSemi\,\cite{Amorim:2016} which maps (un)supervised samples to nodes of a complete graph, with edges weighted by the Euclidean distance between samples. The cost of a path connecting two nodes is the maximal edge-weight on that path. OPFSemi uses this graph to compute an optimum-path forest of minimum-cost paths rooted in the supervised samples. Each supervised sample assigns its label to its most closely connected unsupervised nodes. OPFSemi was shown to perfom better for pseudo-label propagation than earlier semi-supervised methods\,\cite{Amorim:2016,BenatoSibgrapi:2018,Amorim:2019}.

\subsection{Classifier training with pseudo-labels}
\label{sec:classif_training}
To finally test the quality of our generated pseudo-labels, we train a deep neural network, namely VGG-16 with ImageNet pre-trained weights, and test it on our parasite datasets (Fig.~\ref{pipeline_fig1}, bottom box). This architecture was shown to have the best results for our datasets\,\cite{Osaku:2020}. 

\section{\uppercase{Experiments and results}}
\label{sec_experiments}
\subsection{Datasets}
\label{subsec_datasets}
\begin{figure}[htb!]
    \centering
    \includegraphics[width=\linewidth]{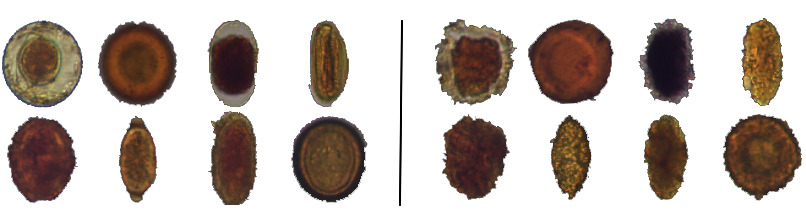}
    \caption{H.eggs dataset. Left: images of parasites of the eight classes in this dataset. Right: Corresponding images of the impurities for each of the left classes which jointly form class 9 (impurities).}
    \label{parasites_fig}
\end{figure}

As outlined in Sec.~\ref{sec:introduction}, we apply our proposed approach in the medical context. Our data (see Tab.~\ref{t.datasets}) consists of five image datasets of Brazil's most common species of human intestinal parasites which are responsible for public health problems and death in infants and immunodeficient individuals in most tropical countries\,\cite{Suzuki:2013}.
The first three datasets contain color microscopy images of 200 $\times$ 200 pixels: (i) \emph{Helminth larvae} (H.larvae, 2 classes, $3,514$ images); (ii) \emph{Helminth eggs} (H.eggs, 9 classes, $5,112$ images, see examples in Fig.~\ref{parasites_fig}); and (iii) \emph{Protozoan cysts} (P.cysts, 7 classes, $9,568$ images). These datasets are unbalanced and they also contain an impurity (adversarial) class that is very similar to the parasite classes, making the problem even more challenging. To evaluate different difficulty levels, we also explore (ii) and (iii) without the impurity class, which form our last two datasets. 

\begin{table}[htb!]
\caption{Parasites datasets. The class names, number of classes, and number of samples per class are presented.}
\label{t.datasets}
\centering
\begin{adjustbox}{width=0.6\linewidth}
\begin{tabular}{l|l|r}
\multicolumn{1}{c|}{\textbf{dataset}}                                                           & \multicolumn{1}{c|}{\textbf{classes}} & \multicolumn{1}{c}{\textbf{$\#$ samples}} \\ \hline
\multirow{3}{*}{\begin{tabular}[c]{@{}l@{}}(i) \textit{H.larvae}\\ (2 classes)\end{tabular}} & \textit{S.stercoralis}                     & 446                              \\
                                                                                       & impurities                 & 3068                              \\
                                                                                       &                           total & 3,514                   \\ \hline
\multirow{10}{*}{\begin{tabular}[c]{@{}l@{}}(ii) \textit{H.eggs}\\ (9 classes)\end{tabular}}  & \textit{H.nana}                     & 348                              \\
                                                                                       & \textit{H.diminuta}                 & 80                              \\
                                                                                       & \textit{Ancilostomideo}             & 148                              \\
                                                                                       & \textit{E.vermicularis}             & 122                              \\
                                                                                       & \textit{A.lumbricoides}             & 337                              \\
                                                                                       & \textit{T.trichiura}                & 375                              \\
                                                                                       & \textit{S.mansoni}                  & 122                              \\
                                                                                       & \textit{Taenia}                     & 236                              \\
                                                                                       & impurities                 & 3,444                              \\
                                                                                       &                          total  & 5,112                   \\ \hline
\multirow{8}{*}{\begin{tabular}[c]{@{}l@{}}(iii) \textit{P.cysts}\\ (7 classes)\end{tabular}} & \textit{E.coli}                     & 719                              \\
                                                                                       & \textit{E.histolytica}              & 78                              \\
                                                                                       & \textit{E.nana}                     & 724                              \\
                                                                                       & \textit{Giardia}                    & 641                              \\
                                                                                       & \textit{I.butschlii}                & 1,501                              \\
                                                                                       & \textit{B.hominis}                  & 189                              \\
                                                                                       & impurities                 & 5,716                              \\
                                                                                       &                           total & 9,568  \\ \hline                       
\end{tabular}
\end{adjustbox}
\end{table}

\subsection{Experimental setup}
\label{subsec_experimental_setup}
As outlined in Sec.~\ref{sec:introduction}, we aim to build a classifier for our image data using a very small set of supervised samples. For this, we split each of the five considered datasets $D$ (Sec.~\ref{subsec_datasets}) into a supervised training-set $S$ containing $1\%$ supervised samples from $D$, an unsupervised training-set $U$ with $69\%$ of the samples in $D$, and a test set $T$ with $30\%$ of the samples in $D$ (hence, $D= S \cup U \cup T$). We repeat the above division randomly and in a stratified manner to create three distinct splits of $D$ in order to gain statistical relevance when evaluating results next.

Table~\ref{t.samples} shows the sizes $|S|$ and $|U|$ for each dataset. To measure quality, we compute accuracy (number of correct classified or labeled samples over all the samples in a set) and Cohen's $\kappa$ (since our datasets are unbalanced). $\kappa$ gives the agreement level between two distinct predictions in a range $[-1,1]$, where $\kappa \leq 0$ means no possibility, and $\kappa = 1$ means full possibility, of agreement. 

\begin{table}[bht!]
\caption{Number of samples in $S$ and $U$ for each dataset.}
\label{t.samples}
\centering

\begin{adjustbox}{width=0.8\linewidth}
\begin{tabular}{c|c|c|c|c|c}
\multicolumn{1}{l|}{} &  \begin{tabular}[c]{@{}c@{}}\textbf{H.eggs}\\ \textbf{(w/o imp)}\end{tabular} & \begin{tabular}[c]{@{}c@{}}\textbf{P. cysts}\\ \textbf{(w/o imp)}\end{tabular} & \textbf{H. larvae} & \textbf{H. eggs} & \textbf{P. cysts} \\ \hline
\textit{S}                                                                            & 17                                                         & 38                                                           & 35        & 51      & 95       \\ \hline
\textit{U}                                                                          & 1220                                                       & 2658                                                         & 2424      & 3527    & 6602    \\ \hline
\end{tabular}
\end{adjustbox}
\end{table}
\subsection{Implementation details}
\label{subsec_implementation_details}
We next outline our end-to-end implementation.

\smallskip
\noindent\textbf{Contrastive learning:}
We implemented SimCLR and SupCon in Python using Pytorch. We generate two augmented images (views) for each original image by random horizontal flip, resized crop ($96\times96$), color jitter (brightness$=0.5$, contrast$=0.5$, saturation$=0.5$, hue$=0.1$) with probability of $0.8$, gray-scale with probability of $0.2$, Gaussian blur ($9\times9$), and a normalization of $0.5$. 

\smallskip
\noindent\textbf{Latent space generation:} 
We replace ResNet-18's decision layer by a linear layer with $4,096$ neurons, a ReLU activation layer, and a linear layer with $1,024$ neurons respectively. We train the model by backpropagating errors of NT-Xent and SupCon losses for SimCLR and SupCon, respectively, with a fixed temperature of $0.07$. We use the AdamW optimizer with a learning rate of $0.0005$, weight decay of $0.0001$, and a learning rate scheduler using cosine annealing, with a maximum temperature equal to the epochs and minimum learning rate of $0.0005/50$. We use $50$ epochs and select the best model through a checkpoint obtained from the lowest validation loss during training. Finally, we use the $512$ features of the ResNet-18's encoder to obtain our latent space.

\smallskip
\noindent\textbf{Classifier using pseudo-labels:} 
We replace the original VGG-16 classifier with two linear layers with $4,096$ neurons followed by ReLU activations and a softmax decision layer. We train the model with the last four layers unfixed by backpropagating errors using categorical cross-entropy. We use stochastic gradient descent with a linear decay learning rate initialized at $0.1$ and momentum of $0.9$ over $15$ epochs.

\smallskip
\noindent\textbf{Parameter setting:}  OPFSup and OPFSemi, used for pseudo-labeling (Sec.~\ref{sec:pseudolabeling}), have no parameters. For Linear SVM and t-SNE (Sec.~\ref{subsubsec_q1}), we use the default parameters provided by scikit-learn.

For replication purposes, all our code and results are made openly available\,\cite{software}.

\subsection{Proposed experiments}
\label{subsec_proposedexp}
To describe our experiments, we first introduce a few notations. $S$, $U$, and $T$ are the supervised (known labels), unsupervised (to be pseudo-labeled), and test sets (see Sec.~\ref{subsec_experimental_setup}). 
Let $I$ be the images in a given dataset having true labels $L$ and pseudo-labels $P$. Let $F$ be the latent features obtained by the three contrastive learning methods; and let $F'$ be the features' projection to 2D via t-SNE. 
We use subscripts to denote on which subset $I$, $L$, $P$, and $F$ are computed, \emph{e.g.} $F_S$ are the latent features for samples in $S$. Finally, let $A$ be the initialization strategy for training a classifier $C$. 

\begin{figure}[htb!]
    \centering
    \includegraphics[width=\linewidth]{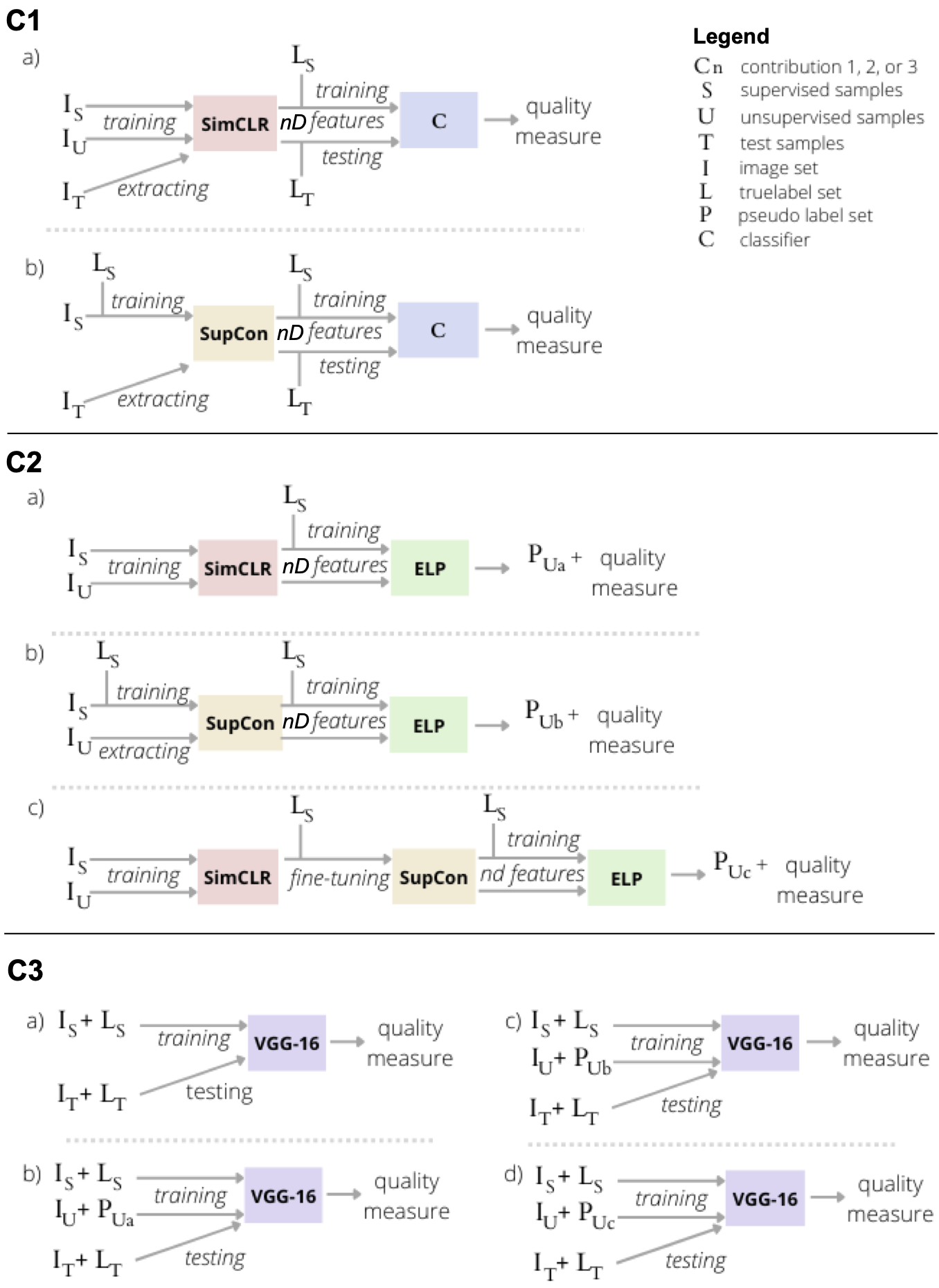}
    \caption{Summary of the proposed experiments.}
    \label{exps_summary}
\end{figure}

Figure~\ref{exps_summary} shows the several experiments we executed to explore the claims C1-C3 listed in Sec.~\ref{sec:introduction}. These experiments are detailed next.

\subsubsection{Testing C1} 
\label{subsubsec_q1}
Our claim C1 is that contrastive learning methods produce high separability of classes (\emph{i.e.}, DS) in the produced feature space. Also, our using of contrastive learning has increased propagation accuracy by up to 20\% \emph{vs} using a simpler method, \emph{i.e.}, generating the latent space via autoencoders\,\cite{BenatoSibgrapi:2018}. Directly measuring DS is hard since the concept of data separability is not uniquely and formally defined (see Sec.~\ref{sec:relatedworks}). As such, we assess DS by a `proxy' method: We train two distinct classifiers $C$, both using $1\%$ supervised samples. These are Linear SVM, a simple linear classifier used to check the linear separability of classes in the latent space; and OPFSup\,\cite{Papa2009}, an Euclidean distance-based classifier. If these classifiers yield high quality, it means that DS is high, and conversely. We measure quality by classifier accuracy and $\kappa$ over correctly classified samples in $T$. 

With the above, we execute two experiments -- one per method of latent space generation (see Sec.~\ref{sec:contr_learn}):

\begin{enumerate}[label=\alph*)]
\item \emph{SimCLR:} Train with $A$ on $I_{S\cup U}$; extract features $F_S$ and $F_T$; train $C$ on $F_S$ and $L_S$; test on $F_T$ and $L_T$.

\item \emph{SupCon:} Train with $A$ on $I_S$ and $L_S$; extract features $F_S$ and $F_T$; train and test as above.
\end{enumerate}

\subsubsection{Testing C2} 
\label{subsubsec_q2}
Similarly to C1, evaluating the VS of projections to test C2 can be done in many ways since visual separation of clusters in a 2D scatterplot is a broad concept. In DR literature, several metrics have been proposed for this task (see surveys \,\cite{espadoto19,nonato18}). Yet, such metrics are typically used to gauge the projection quality when explored by a \emph{human}. Rather, in our context, we use projections \emph{automatically} to drive pseudo-labeling and improve classification (Sec.~\ref{sec:pseudolabeling}). As such, it makes sense to evaluate our projections' VS by how well they can do this label propagation. For this, we compare the computed pseudo-labels with the true, supervised, labels by computing accuracy and $\kappa$ for the correctly computed pseudo-labels over $U$. We do this via three experiments:

\begin{enumerate}[label=\alph*)]
    \item \emph{SimCLR:} Train with $A$ on $I_{S\cup U}$; extract features $F_{S\cup U}$; compute 2D features $F'$ with t-SNE from $F_{S\cup U}$; propagate labels $L_S$ with OPFSemi from $F'_S$ to $F'_U$;
    \item \emph{SupCon:} Train with $A$ on $I_S$ and $L_S$; extract features $F_{S\cup U}$; compute 2D features $F'$ with t-SNE from $F_{S\cup U}$; propagate labels as above;
    \item \emph{SimCLR+SupCon:} Train SimCLR with $A$ on $I_{S\cup U}$; fine-tune with SupCon on $I_S$ and $L_S$; extract features $I_{S\cup U}$; compute 2D features $F'$ with t-SNE from $F_{S\cup U}$; propagate labels as above.
\end{enumerate}

\subsubsection{Experiments for testing C3} 
\label{subsubsec_q3}
Finally, we use the computed pseudo-labels to train and test a DNN classifier, namely VGG-16, to test C3, \emph{i.e.}, gauge how CS is correlated (or not) with VS and DS. For this, we do the following experiments:

\begin{enumerate}[label=\alph*)]
    \item \emph{baseline:} train with $I_S$ and $L_S$; test on $I_T$ and $L_T$;
    \item \emph{SimCLR:} train with $I_{S\cup U}$ and $L_{S\cup P_{U}}$, with pseudo-labels $P_U$ from (Sec.~\ref{subsubsec_q2},a); test as above;
    \item \emph{SupCon:} train with $I_{S\cup U}$ and $L_{S\cup P_{U}}$, with pseudo-labels $P_U$ from (Sec.~\ref{subsubsec_q2},b); test as above;
    \item \emph{SimCLR+SupCon:} train with $I_{S\cup U}$ and $L_{S\cup P_{U}}$, with pseudo-labels $P_U$ from (Sec.~\ref{subsubsec_q2},c); test as above.
\end{enumerate}

\subsection{Results}
\label{sec:results}
We present the results of the experiments in Sec.~\ref{subsec_proposedexp} and along our  claims C1-C3.

\subsubsection{C1: Contrastive learning yields high DS}

\begin{table*}[ht]
\caption{C1: DS assessment of SimCLR's and SupCon's latent spaces using Linear SVM and OPFSup on $T$. Both methods are compared trained from scratch and with pre-trained weights during 50 epochs. Best values per dataset are in bold.}
\label{exps_table1}
\centering
\begin{adjustbox}{width=\textwidth}

\begin{tabular}{ll|llll|llll|}
\cline{3-10}
                                                    &       & \multicolumn{4}{c|}{ \textbf{trained from scratch}}                                                                                                                                     & \multicolumn{4}{c|}{ \textbf{with ImageNet pre-trained weights}}                                                                                                                                               \\ \cline{3-10} 
                                                    &       & \multicolumn{2}{c|}{a) SimCLR}                                                           & \multicolumn{2}{c|}{b) SupCon}                                                   & \multicolumn{2}{c|}{a) SimCLR}                                                                    & \multicolumn{2}{c|}{b) SupCon}                                                   \\ \cline{3-10} 
                                                    &       & \multicolumn{1}{c|}{Linear SVM}                  & \multicolumn{1}{c|}{OPFSup}                  & \multicolumn{1}{c|}{Linear SVM}                           & \multicolumn{1}{c|}{OPFSup} & \multicolumn{1}{c|}{Linear SVM}                           & \multicolumn{1}{c|}{OPFSup}                  & \multicolumn{1}{c|}{Linear SVM}                           & \multicolumn{1}{c|}{OPFSup} \\ \hline
\multicolumn{1}{|l|}{\multirow{2}{*}{\begin{tabular}[c]{@{}l@{}}H.eggs\\ (w/o imp)\end{tabular}}}   & acc   & \multicolumn{1}{l|}{0.814606 $\pm$ 0.079} & \multicolumn{1}{l|}{0.759631 $\pm$ 0.107} & \multicolumn{1}{l|}{0.863954 $\pm$ 0.064}          & 0.858565 $\pm$ 0.057     & \multicolumn{1}{l|}{\textbf{0.903327 $\pm$ 0.021}} & \multicolumn{1}{l|}{0.869429 $\pm$ 0.033} & \multicolumn{1}{l|}{0.789705 $\pm$ 0.042}          & 0.817326 $\pm$ 0.047     \\ \cline{2-10} 
\multicolumn{1}{|l|}{}                              & $\kappa$ & \multicolumn{1}{l|}{0.668252 $\pm$ 0.091} & \multicolumn{1}{l|}{0.585225 $\pm$ 0.098} & \multicolumn{1}{l|}{0.778473 $\pm$ 0.029}          & 0.742304 $\pm$ 0.106     & \multicolumn{1}{l|}{\textbf{0.884889 $\pm$ 0.025}} & \multicolumn{1}{l|}{0.844527 $\pm$ 0.04}  & \multicolumn{1}{l|}{0.750924 $\pm$ 0.049}          & 0.783428 $\pm$ 0.056     \\ \hline
\multicolumn{1}{|l|}{\multirow{2}{*}{\begin{tabular}[c]{@{}l@{}}P.cysts\\ (w/o imp)\end{tabular}}}  & acc   & \multicolumn{1}{l|}{0.637543 $\pm$ 0.177} & \multicolumn{1}{l|}{0.632065 $\pm$ 0.017} & \multicolumn{1}{l|}{0.717705 $\pm$ 0.022}          & 0.643310 $\pm$ 0.045      & \multicolumn{1}{l|}{\textbf{0.771627 $\pm$ 0.019}} & \multicolumn{1}{l|}{0.706747 $\pm$ 0.038} & \multicolumn{1}{l|}{0.675606 $\pm$ 0.056}          & 0.580450 $\pm$ 0.006      \\ \cline{2-10} 
\multicolumn{1}{|l|}{}                              & $\kappa$ & \multicolumn{1}{l|}{0.547758 $\pm$ 0.168} & \multicolumn{1}{l|}{0.523332 $\pm$ 0.02}  & \multicolumn{1}{l|}{0.615566 $\pm$ 0.025}          & 0.529749 $\pm$ 0.053     & \multicolumn{1}{l|}{\textbf{0.689346 $\pm$ 0.027}} & \multicolumn{1}{l|}{0.605509 $\pm$ 0.049} & \multicolumn{1}{l|}{0.564481 $\pm$ 0.061}          & 0.443273 $\pm$ 0.015     \\ \hline
\multicolumn{1}{|l|}{\multirow{2}{*}{H.larvae}} & acc   & \multicolumn{1}{l|}{0.901106 $\pm$ 0.025} & \multicolumn{1}{l|}{0.888784 $\pm$ 0.011} & \multicolumn{1}{l|}{0.933649 $\pm$ 0.011}          & 0.905845 $\pm$ 0.033     & \multicolumn{1}{l|}{0.950079 $\pm$ 0.006}          & \multicolumn{1}{l|}{0.947551 $\pm$ 0.008} & \multicolumn{1}{l|}{\textbf{0.952923 $\pm$ 0.007}} & 0.946287 $\pm$ 0.008     \\ \cline{2-10} 
\multicolumn{1}{|l|}{}                              & $\kappa$ & \multicolumn{1}{l|}{0.381798 $\pm$ 0.233} & \multicolumn{1}{l|}{0.422084 $\pm$ 0.037} & \multicolumn{1}{l|}{0.711252 $\pm$ 0.069}          & 0.539386 $\pm$ 0.237     & \multicolumn{1}{l|}{0.767091 $\pm$ 0.041}          & \multicolumn{1}{l|}{0.751936 $\pm$ 0.054} & \multicolumn{1}{l|}{\textbf{0.782983 $\pm$ 0.053}} & 0.756410 $\pm$ 0.054      \\ \hline
\multicolumn{1}{|l|}{\multirow{2}{*}{H.eggs}}   & acc   & \multicolumn{1}{l|}{0.542590 $\pm$ 0.177}  & \multicolumn{1}{l|}{0.575185 $\pm$ 0.014} & \multicolumn{1}{l|}{\textbf{0.789222 $\pm$ 0.028}} & 0.756410 $\pm$ 0.035      & \multicolumn{1}{l|}{0.758800 $\pm$ 0.053}            & \multicolumn{1}{l|}{0.736202 $\pm$ 0.029} & \multicolumn{1}{l|}{0.761191 $\pm$ 0.071}          & 0.743590 $\pm$ 0.069      \\ \cline{2-10} 
\multicolumn{1}{|l|}{}                              & $\kappa$ & \multicolumn{1}{l|}{0.126531 $\pm$ 0.046} & \multicolumn{1}{l|}{0.279272 $\pm$ 0.023} & \multicolumn{1}{l|}{\textbf{0.626696 $\pm$ 0.037}} & 0.592371 $\pm$ 0.039     & \multicolumn{1}{l|}{0.529617 $\pm$ 0.125}          & \multicolumn{1}{l|}{0.521839 $\pm$ 0.056} & \multicolumn{1}{l|}{0.588783 $\pm$ 0.111}          & 0.567762 $\pm$ 0.095     \\ \hline
\multicolumn{1}{|l|}{\multirow{2}{*}{P.cysts}}  & acc   & \multicolumn{1}{l|}{0.563335 $\pm$ 0.045} & \multicolumn{1}{l|}{0.541159 $\pm$ 0.018} & \multicolumn{1}{l|}{\textbf{0.722048 $\pm$ 0.009}} & 0.609544 $\pm$ 0.019     & \multicolumn{1}{l|}{0.674678 $\pm$ 0.064}          & \multicolumn{1}{l|}{0.604551 $\pm$ 0.023} & \multicolumn{1}{l|}{0.628701 $\pm$ 0.168}          & 0.649483 $\pm$ 0.05      \\ \cline{2-10} 
\multicolumn{1}{|l|}{}                              & $\kappa$ & \multicolumn{1}{l|}{0.330526 $\pm$ 0.031} & \multicolumn{1}{l|}{0.288527 $\pm$ 0.012} & \multicolumn{1}{l|}{\textbf{0.525391 $\pm$ 0.045}} & 0.370582 $\pm$ 0.022     & \multicolumn{1}{l|}{0.422320 $\pm$ 0.112}           & \multicolumn{1}{l|}{0.375311 $\pm$ 0.037} & \multicolumn{1}{l|}{0.441970 $\pm$ 0.168}           & 0.429321 $\pm$ 0.065     \\ \hline
\end{tabular}

\end{adjustbox}
\end{table*}

Table~\ref{exps_table1} shows the classification results for the experiments in Sec.~\ref{subsubsec_q1} in terms of accuracy and $\kappa$ (mean and standard deviation) for the trained Linear SVM and OPFSup classifiers.

We first discuss the contrastive learning methods trained from scratch \emph{vs} using ImageNet pre-trained weights. For all datasets, the best accuracy and $\kappa$ exceed $0.70$ and $0.50$ respectively. Linear SVM obtained the best results, showing that the tested latent spaces have a reasonable \emph{linear separation} between classes even when classified with only $1\%$ supervised samples. In contrast, OPFSup seems to suffer from the dimensionality curse as it uses Euclidean distances in the latent space. This further motivates the latent space's dimensionality reduction when using an OPF classifier. Separately, we see that SimCLR was helped by the ImageNet pre-trained weights, while SupCon obtained its best results when trained from scratch for datasets with impurities. SimCLR had an increase of around $0.10$ in accuracy and $\kappa$ for H.eggs and P.cysts without impurities with pre-trained weights. SupCon also had an extra $0.10$ accuracy and $\kappa$ for datasets with impurities when trained from scratch. Since SupCon achieved its best results from scratch and SimCLR was helped by pre-trained weights for distinct datasets, we next explore the combination of both methods.

\subsubsection{C2: Projections of contrastive latent spaces yield high VS}
Table~\ref{exps_table2} show the results for the experiments in Sec.~\ref{subsubsec_q2}, \emph{i.e.}, the mean propagation accuracy and $\kappa$ in pseudo-labeling for the correctly assigned labels in $U$ for EPL run on latent spaces created by SimCLR, SupCon, and SimCLR+SupCon. 

The best results were obtained when using the ImageNet pre-trained weights. This shows that the pseudo-labeling on the contrastive latent space is favored by such pre-trained weights. SupCon gained almost $0.20$ in $\kappa$ compared with SimCLR for H.eggs and P.cysts without impurity. SupCon obtained the best results for the H.Eggs and P.cysts without impurities, while the SimCLR+SupCon obtained the best results for the same datasets with impurities. SimCLR+SupCon improved the results of SimCLR for those datasets. For H.larvae, the results of the three methods were similar.

\begin{table*}[ht]
\centering
\caption{C2: Propagation results for pseudo-labeling $U$ on the projected SimCLR's and SupCon's latent spaces, from scratch and using ImageNet pre-trained weights. Best values per dataset are in bold.}
\label{exps_table2}
\begin{adjustbox}{width=\textwidth}

\begin{tabular}{ll|rrr|rrr|}
\cline{3-8}
                                                    &             & \multicolumn{3}{c|}{\textbf{trained from scratch}}                                                                                          & \multicolumn{3}{c|}{\textbf{with ImageNet pre-trained weights}}                                                                                                             \\ \cline{3-8} 
                                                    &             & \multicolumn{1}{c|}{a) SimCLR}               & \multicolumn{1}{c|}{b) SupCon}               & \multicolumn{1}{c|}{c) SimCLR+SupCon} & \multicolumn{1}{c|}{a) SimCLR}                        & \multicolumn{1}{c|}{b) SupCon}                        & \multicolumn{1}{c|}{c) SimCLR+SupCon} \\ \hline
\multicolumn{1}{|l|}{\multirow{2}{*}{\begin{tabular}[c]{@{}l@{}}H.eggs\\ (w/o imp)\end{tabular}}}   &   acc   & \multicolumn{1}{r|}{0.861493 $\pm$ 0.012} & \multicolumn{1}{r|}{0.713554 $\pm$ 0.077} & 0.896255 $\pm$ 0.041               & \multicolumn{1}{r|}{0.795203 $\pm$ 0.129}          & \multicolumn{1}{r|}{\textbf{0.951765 $\pm$ 0.041}} & 0.830234 $\pm$ 0.123               \\ \cline{2-8} 
\multicolumn{1}{|l|}{}                              &   $\kappa$ & \multicolumn{1}{r|}{0.561568 $\pm$ 0.009} & \multicolumn{1}{r|}{0.473379 $\pm$ 0.025} & 0.567093 $\pm$ 0.020                & \multicolumn{1}{r|}{0.756312 $\pm$ 0.153}          & \multicolumn{1}{r|}{\textbf{0.942519 $\pm$ 0.049}} & 0.797482 $\pm$ 0.148               \\ \hline
\multicolumn{1}{|l|}{\multirow{2}{*}{\begin{tabular}[c]{@{}l@{}}P.cysts\\ (w/o imp)\end{tabular}}}  &   acc   & \multicolumn{1}{r|}{0.652324 $\pm$ 0.027} & \multicolumn{1}{r|}{0.641073 $\pm$ 0.038} & 0.650470 $\pm$ 0.027                & \multicolumn{1}{r|}{0.568991 $\pm$ 0.036}          & \multicolumn{1}{r|}{\textbf{0.706973 $\pm$ 0.092}} & 0.565282 $\pm$ 0.091               \\ \cline{2-8} 
\multicolumn{1}{|l|}{}                              &   $\kappa$ & \multicolumn{1}{r|}{0.537704 $\pm$ 0.043} & \multicolumn{1}{r|}{0.531090 $\pm$ 0.040}   & 0.533208 $\pm$ 0.031               & \multicolumn{1}{r|}{0.428962 $\pm$ 0.036}          & \multicolumn{1}{r|}{\textbf{0.619738 $\pm$ 0.102}} & 0.439581 $\pm$ 0.103               \\ \hline
\multicolumn{1}{|l|}{\multirow{2}{*}{H.larvae}} &   acc   & \multicolumn{1}{r|}{0.898739 $\pm$ 0.033} & \multicolumn{1}{r|}{0.886539 $\pm$ 0.003} & 0.941169 $\pm$ 0.013               & \multicolumn{1}{r|}{\textbf{0.959062 $\pm$ 0.007}} & \multicolumn{1}{r|}{0.946184 $\pm$ 0.010}           & 0.954724 $\pm$ 0.005               \\ \cline{2-8} 
\multicolumn{1}{|l|}{}                              &   $\kappa$ & \multicolumn{1}{r|}{0.532710 $\pm$ 0.179}  & \multicolumn{1}{r|}{0.404983 $\pm$ 0.119} & 0.694591 $\pm$ 0.029               & \multicolumn{1}{r|}{\textbf{0.817274 $\pm$ 0.030}}  & \multicolumn{1}{r|}{0.777621 $\pm$ 0.020}           & 0.792838 $\pm$ 0.009               \\ \hline
\multicolumn{1}{|l|}{\multirow{2}{*}{H.eggs}}   &   acc   & \multicolumn{1}{r|}{0.710173 $\pm$ 0.035} & \multicolumn{1}{r|}{0.585802 $\pm$ 0.026} & 0.741755 $\pm$ 0.065               & \multicolumn{1}{r|}{0.719862 $\pm$ 0.077}          & \multicolumn{1}{r|}{0.751723 $\pm$ 0.052}          & \textbf{0.780418 $\pm$ 0.080}       \\ \cline{2-8} 
\multicolumn{1}{|l|}{}                              &   $\kappa$ & \multicolumn{1}{r|}{0.357514 $\pm$ 0.044} & \multicolumn{1}{r|}{0.178536 $\pm$ 0.031} & 0.374099 $\pm$ 0.108               & \multicolumn{1}{r|}{0.532788 $\pm$ 0.120}           & \multicolumn{1}{r|}{0.553654 $\pm$ 0.094}          & \textbf{0.624724 $\pm$ 0.113}      \\ \hline
\multicolumn{1}{|l|}{\multirow{2}{*}{P.cysts}}  &   acc   & \multicolumn{1}{r|}{0.607884 $\pm$ 0.049} & \multicolumn{1}{r|}{0.530785 $\pm$ 0.019} & 0.666119 $\pm$ 0.027               & \multicolumn{1}{r|}{0.670898 $\pm$ 0.051}          & \multicolumn{1}{r|}{0.577025 $\pm$ 0.049}          & \textbf{0.705042 $\pm$ 0.035}      \\ \cline{2-8} 
\multicolumn{1}{|l|}{}                              &   $\kappa$ & \multicolumn{1}{r|}{0.380969 $\pm$ 0.066} & \multicolumn{1}{r|}{0.235849 $\pm$ 0.018} & 0.457391 $\pm$ 0.056               & \multicolumn{1}{r|}{0.430201 $\pm$ 0.022}          & \multicolumn{1}{r|}{0.320479 $\pm$ 0.057}          & \textbf{0.513962 $\pm$ 0.043}      \\ \hline
\end{tabular}

\end{adjustbox}
\end{table*}

\subsubsection{C3: Classifiers trained by pseudo-labels obtained from high-VS projections have a high CP}
Table~\ref{exps_table3} shows the results of classification for VGG-16 trained from the pseudo-labeling performed on latent spaces from SimCLR, SupCon, and SimCLR+SupCon. 

We notice that the results of VGG-16's classification follow the same pattern as the propagation results (Tab.~\ref{exps_table2}). The best results were found by the methods using the ImageNet pre-trained weights. Also, SupCon obtained the best results for H.Eggs and P.cysts without impurities, while SimCLR+SupCon obtained the best results for the same datasets with impurities. SupCon showed a gain of almost $0.20$ in $\kappa$ for H.eggs without impurity and H.larvae, and $0.15$ for P.cysts without impurity when compared with the baseline. In short, the results show that VGG-16 can learn from the pseudo-labels since it provided good classification accuracies and $\kappa$ -- higher than $0.85$ and $0.76$, respectively -- for H.eggs and P.cysts without impurity and H.Larvae. However, the compared methods could not surpass the baseline for H.eggs and P.cysts with impurities. We discuss this aspect next. 

\begin{table*}[htbp!]
\centering
\caption{C3: VGG-16's classification results on $T$ when using pseudo labels from SimCLR's, SupCon and SimCLR+SupCon latent spaces, from scratch and with ImageNet pre-trained weights. Best values per dataset are in bold.}
\label{exps_table3}
\begin{adjustbox}{width=\textwidth}

\begin{tabular}{ll|r|rrr|rrr|}
\cline{3-9}
                                                                                                   &       & \multicolumn{1}{c|}{\multirow{2}{*}{a) baseline}} & \multicolumn{3}{c|}{\textbf{trained from scratch}}                                                                                      & \multicolumn{3}{c|}{\textbf{with ImageNet pre-trained weights}}                                                                                                         \\ \cline{4-9} 
                                                                                                   &       & \multicolumn{1}{c|}{}                          & \multicolumn{1}{c|}{b) SimCLR}               & \multicolumn{1}{c|}{c) SupCon}               & \multicolumn{1}{c|}{d) SimCLR+SupCon} & \multicolumn{1}{c|}{a) SimCLR}                        & \multicolumn{1}{c|}{b) SupCon}                        & \multicolumn{1}{c|}{c) SimCLR+SupCon} \\ \hline
\multicolumn{1}{|l|}{\multirow{2}{*}{\begin{tabular}[c]{@{}l@{}}H.eggs\\ (w/o imp)\end{tabular}}}  & acc   & 0.812932 $\pm$ 0.059                           & \multicolumn{1}{r|}{0.435028 $\pm$ 0.400}   & \multicolumn{1}{r|}{0.714375 $\pm$ 0.088} & 0.925926 $\pm$ 0.035               & \multicolumn{1}{r|}{0.823603 $\pm$ 0.138}          & \multicolumn{1}{r|}{\textbf{0.961080 $\pm$ 0.039}}  & 0.858129 $\pm$ 0.127               \\ \cline{2-9} 
\multicolumn{1}{|l|}{}                                                                             & $\kappa$ & 0.775954 $\pm$ 0.073                           & \multicolumn{1}{r|}{0.292310 $\pm$ 0.506}  & \multicolumn{1}{r|}{0.662603 $\pm$ 0.098} & 0.912482 $\pm$ 0.041               & \multicolumn{1}{r|}{0.790296 $\pm$ 0.164}          & \multicolumn{1}{r|}{\textbf{0.953710 $\pm$ 0.047}}  & 0.831107 $\pm$ 0.152               \\ \hline
\multicolumn{1}{|l|}{\multirow{2}{*}{\begin{tabular}[c]{@{}l@{}}P.cysts\\ (w/o imp)\end{tabular}}} & acc   & 0.757209 $\pm$ 0.015                           & \multicolumn{1}{r|}{0.589965 $\pm$ 0.174} & \multicolumn{1}{r|}{0.662053 $\pm$ 0.064} & 0.606113 $\pm$ 0.188               & \multicolumn{1}{r|}{0.752905 $\pm$ 0.183}          & \multicolumn{1}{r|}{\textbf{0.857411 $\pm$ 0.085}} & 0.740945 $\pm$ 0.216               \\ \cline{2-9} 
\multicolumn{1}{|l|}{}                                                                             & $\kappa$ & 0.651933 $\pm$ 0.023                           & \multicolumn{1}{r|}{0.383736 $\pm$ 0.334} & \multicolumn{1}{r|}{0.558104 $\pm$ 0.071} & 0.408416 $\pm$ 0.354               & \multicolumn{1}{r|}{0.622887 $\pm$ 0.185}          & \multicolumn{1}{r|}{\textbf{0.766192 $\pm$ 0.043}} & 0.608864 $\pm$ 0.200                 \\ \hline
\multicolumn{1}{|l|}{\multirow{2}{*}{H.larvae}}                                                    & acc   & 0.930806 $\pm$ 0.026                           & \multicolumn{1}{r|}{0.903950 $\pm$ 0.034}  & \multicolumn{1}{r|}{0.888784 $\pm$ 0.009} & 0.942496 $\pm$ 0.015               & \multicolumn{1}{r|}{\textbf{0.956714 $\pm$ 0.004}} & \multicolumn{1}{r|}{\textbf{0.952607 $\pm$ 0.008}} & \textbf{0.957978 $\pm$ 0.001}      \\ \cline{2-9} 
\multicolumn{1}{|l|}{}                                                                             & $\kappa$ & 0.613432 $\pm$ 0.233                           & \multicolumn{1}{r|}{0.538558 $\pm$ 0.196} & \multicolumn{1}{r|}{0.406452 $\pm$ 0.168} & 0.738656 $\pm$ 0.061               & \multicolumn{1}{r|}{\textbf{0.809830 $\pm$ 0.019}}  & \multicolumn{1}{r|}{\textbf{0.803148 $\pm$ 0.018}} & \textbf{0.807574 $\pm$ 0.021}      \\ \hline
\multicolumn{1}{|l|}{\multirow{2}{*}{H.eggs}}                                                      & acc   & \textbf{0.862234 $\pm$ 0.015}                  & \multicolumn{1}{r|}{0.728814 $\pm$ 0.059} & \multicolumn{1}{r|}{0.606693 $\pm$ 0.042} & 0.779444 $\pm$ 0.073               & \multicolumn{1}{r|}{0.737723 $\pm$ 0.068}          & \multicolumn{1}{r|}{0.780095 $\pm$ 0.060}           & 0.806389 $\pm$ 0.073               \\ \cline{2-9} 
\multicolumn{1}{|l|}{}                                                                             & $\kappa$ & \textbf{0.740861 $\pm$ 0.028}                  & \multicolumn{1}{r|}{0.566056 $\pm$ 0.064} & \multicolumn{1}{r|}{0.286646 $\pm$ 0.063} & 0.627849 $\pm$ 0.099               & \multicolumn{1}{r|}{0.553855 $\pm$ 0.114}          & \multicolumn{1}{r|}{0.592800 $\pm$ 0.116}            & 0.661330 $\pm$ 0.103                \\ \hline
\multicolumn{1}{|l|}{\multirow{2}{*}{P.cysts}}                                                     & acc   & \textbf{0.850691 $\pm$ 0.018}                  & \multicolumn{1}{r|}{0.687333 $\pm$ 0.028} & \multicolumn{1}{r|}{0.379775 $\pm$ 0.020}  & 0.703820 $\pm$ 0.020                 & \multicolumn{1}{r|}{0.725648 $\pm$ 0.036}          & \multicolumn{1}{r|}{0.645304 $\pm$ 0.052}          & 0.737258 $\pm$ 0.036               \\ \cline{2-9} 
\multicolumn{1}{|l|}{}                                                                             & $\kappa$ & \textbf{0.751667 $\pm$ 0.028}                  & \multicolumn{1}{r|}{0.429244 $\pm$ 0.179} & \multicolumn{1}{r|}{0.184170 $\pm$ 0.023}  & 0.522443 $\pm$ 0.027               & \multicolumn{1}{r|}{0.540847 $\pm$ 0.049}          & \multicolumn{1}{r|}{0.395300 $\pm$ 0.079}            & 0.565966 $\pm$ 0.045               \\ \hline
\end{tabular}

\end{adjustbox}
\end{table*}

\section{\uppercase{Discussion}}

We next discuss several aspects pertaining to our results.

\subsection{Visual separation \emph{vs} classifier performance}

\begin{figure*}[htb!]
    \centering
    \includegraphics[width=\linewidth]{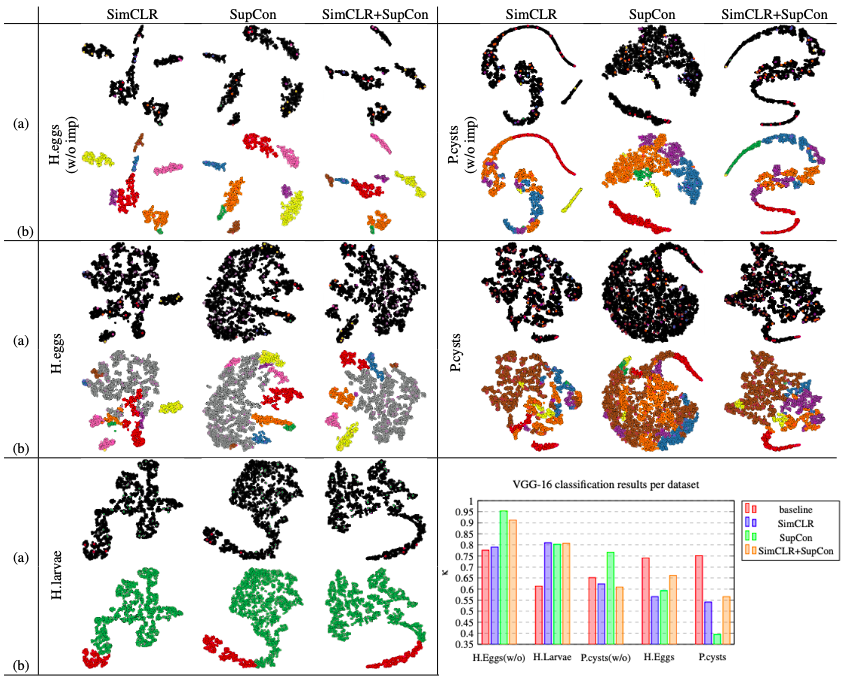}
    \caption{(i) top left: t-SNE projections of the three contrastive latent spaces (SimCLR, SupCon, SimCLR+SupCon) for the six studied datasets. In (a), black points are the unsupervised samples $U$ and colored points the supervised ones $S$. In (b), colors show the computed pseudo-labels. (ii) bottom right: $\kappa$ values for baseline, SimCLR, SupCon, and SimCLR+SupCon experiments. Datasets are ordered on higher $\kappa$ values from left to right.}
    \label{fig_projectionsnew}
\end{figure*}

Figure~\ref{fig_projectionsnew}.i shows the 2D t-SNE projections of the three computed latent spaces for all five studied datasets. For each dataset, the top row (a) shows the few (1\%) supervised labels (colored points) thinly spread among the vast majority of unsupervised (black) samples; the bottom row (b) shows samples colored by the computed pseudo-labels.

We see in all images a good correlation of the visual separation VS (point groups separated from each other by whitespace) with the lack of label mixing in such groups. For H.eggs without impurity, all three latent space projections show a clear VS, and we see that this leads to almost no color mixing in the propagated pseudo-labels. For the H.eggs dataset, we see how the visually separated groups show almost no color mixing, whereas the parts of the projection where no VS is present show color mixing. For P.cysts without impurity, there is a clearly separated group at the bottom in all three projections which also has a single color (label). The remaining parts of the projections, which have no clear VS into distinct groups, show a mix of different colors. For P.cysts, the projections have even less VS, and we see how labels get even more mixed -- for instance, the impurity class (brown) is spread all over the projection. For H.larvae, the larvae class (red) is better separated from the big group of impurities (green), and this correlates with the larvae samples being all located in a tail-like periphery of the projection -- thus, better visually separated from the rest.

All in all, these results show that a good VS leads to a low mixing of the propagated labels, and conversely. In turn, a low mixing will lead to a high classification performance (CP), and conversely, \emph{i.e.}, our claim C3. Figure~\ref{fig_projectionsnew}.ii shows this by comparing the results for the baseline and for VGG-16 trained with the generated pseudo-labels. We see a gain of almost $0.20$ in $\kappa$ from baseline (red) to the proposed pseudo-labeling method (green) for those datasets with a clear VS and little label mixing in the projections. Conversely, we see the CP results are are below to baseline for the datasets with poor VS and color-mixing in their projections.

\subsection{Contrastive learning from few supervised samples}
Our experiments show that SimCLR -- even trained with \emph{thousands} of unsupervised samples (69\%) -- and having more information on the data distribution of the original space -- could not overpass SupCon which used only \emph{dozens} of supervised samples (1\%). The only explanation we find for this is that the latent space generated when SupCon was used to fine-tune SimCLR (SupCon+SimCLR) had a \emph{better data separation} (DS) than the one created by SimCLR. On the one hand, this shows the benefit of using SupCon with supervised data restriction as compared to SimCLR, a comparison that up to our knowledge has not been done before. On the other hand, having a higher DS lead to a higher CP further supports our claim C3.



\section{\uppercase{Conclusion}}
In this paper, we proposed a method to create high-quality classifiers for image datasets from training-sets having only very few supervised (labeled) samples. For this, we used two contrastive learning approaches (SimCLR and SupCon) as well as a combination of the two to generate latent spaces. Next, we projected these spaces to 2D using t-SNE, propagated labels in the projection, and finally used these pseudo-labels to train a final deep-learning classifier for a challenging problem involving the classification of human intestinal parasite images.

Our results show that SupCon performed better than SimCLR when only $1\%$ of supervised samples were available, even though SimCLR uses thousands of distinct samples of the data distribution. We showed label propagation accuracies up to 95\% for the studied datasets without impurities (an adversarial class) and up to 70\% for datasets with impurities, respectively.

Additionally, our experiments show that a high data separation (DS) in the latent space leads to a high visual separation (VS) in the 2D projection which, in turn, leads to high classifier performance (CP). While partial results of this kind have been presented by earlier infovis and machine learning papers, our work is, to our knowledge, the first time that DS, VS, and CP are all linked in the context of an application involving the generation of rich training-sets by pseudo-labeling.

Several future work directions are possible. First, the VS-CP correlation directly suggests that it is interesting to explore using different projection methods than t-SNE. If such methods lead to a higher VS for a given DS, then they will very likely lead to a higher final CP, thus, better classifiers. Secondly, we aim to involve users in the loop to assist the automatic pseudo-labeling process by \emph{e.g.} adjusting some of the automatically propagated labels based on the human assessment of VS. We believe that this will lead to even more accurate pseudo-labels and, ultimately, more accurate classifiers for the problem at hand.

 

\section*{\uppercase{Acknowledgments}}
The authors acknowledge FAPESP grants \#2014/12236-1, \#2019/10705-8, \#2022/12668-5, CAPES grants with Finance Code 001, and CNPq grants \#303808/2018-7.

\bibliographystyle{apalike}
{\small
\bibliography{example}}

\end{document}